\newif\ifeusipstyle
\newif\ifdohybrid
\dohybridfalse

\documentclass[journal]{IEEEtran}

\usepackage[dvips]{graphicx}
\usepackage{url}
\usepackage{amsfonts}
\usepackage{verbatim}
\usepackage{color}
\usepackage{tablefootnote}
\usepackage{acronym}
\usepackage{booktabs,amsmath,tabularx,multirow,adjustbox}
\acrodef{cnn}[CNN]{convolutional neural network}
\acrodef{dnn}[DNN]{deep neural network}
\acrodef{hmm}[HMM]{hidden Markov model}
\acrodef{pbn}[PBN]{projected belief network}
\acrodef{pbn-da}[PBN-DA]{discriminative alignment of \acl{pbn}}
\acrodef{pdf}[PDF]{probability density function}

\title{A Comparison of PDF Projection with Normalizing Flows and SurVAE}
\author{\IEEEauthorblockN{Paul M. Baggenstoss}
   \IEEEauthorblockA{Fraunhofer FKIE,
   Fraunhoferstrasse 20\\
   53343 Wachtberg, Germany\\
   Email: p.m.baggenstoss@ieee.org\\}
   \and
   \IEEEauthorblockN{Felix Govaers}
   \IEEEauthorblockA{Fraunhofer FKIE\\
   Email: felix.govaers@fkie.fraunhofer.de\\}
   \thanks{This work was supported jointly by the Office of Naval Research Global
   and the Defense Advanced Research Projects Agency under Research Grant - N62909-21-1-2024}
}

\begin{document}
\newcommand{\defined}{\stackrel{\mbox{\tiny$\Delta$}}{=}}
\newtheorem{example}{Example}
\newtheorem{conclusion}{Conclusion}
\newtheorem{assumption}{Assumption}
\newtheorem{definition}{Definition}
\newtheorem{problem}{Problem}
\newcommand{\erf}{{\rm erf}}

\newcommand{\sst}{\scriptstyle }
\newcommand{\xparen}{\mbox{\small$(\bfx)$}}
\newcommand{\hojz}{H_{0j}\mbox{\small$(\bfz)$}}
\newcommand{\Hozj}{H_{0,j}\mbox{\small$(\bfz_j)$}}
\newcommand{\smallmath}[1]{{\scriptstyle #1}}
\newcommand{\Hoz}[1]{H_0\mbox{\small$(#1)$}}
\newcommand{\Hozp}[1]{H_0^\prime\mbox{\small$(#1)$}}
\newcommand{\Hozpp}[1]{H_0^{\prime\prime}\mbox{\small$(#1)$}}
\newcommand{\hoz}{\Hoz{\bfz}}
\newcommand{\hooz}{\Hozp{\bfz}}
\newcommand{\hoooz}{\Hozpp{\bfz}}
\newcommand{\smJ}{{\scriptscriptstyle \! J}}
\newcommand{\smK}{{\scriptscriptstyle \! K}}

\newcommand{\erfc}{{\rm erfc}}
\newcommand{\bitem}{\begin{itemize}}
\newcommand{\dsum}{{ \displaystyle \sum}}
\newcommand{\eitem}{\end{itemize}}
\newcommand{\benum}{\begin{enumerate}}
\newcommand{\eenum}{\end{enumerate}}
\newcommand{\bdm}{\begin{displaymath}}
\newcommand{\bfzro}{{\underline{\bf 0}}}
\newcommand{\bfone}{{\underline{\bf 1}}}
\newcommand{\edm}{\end{displaymath}}
\newcommand{\beq}{\begin{equation}}
\newcommand{\bea}{\begin{eqnarray}}
\newcommand{\eea}{\end{eqnarray}}
\newcommand{\cali}{ {\cal \bf I}}
\newcommand{\caln}{ {\cal \bf N}}
\newcommand{\barray}{\begin{displaymath} \begin{array}{rcl}}
\newcommand{\earray}{\end{array}\end{displaymath}}
\newcommand{\eeq}{\end{equation}}
\newcommand{\qed}{\framebox{$\;$}}
\newcommand{\btheta}{\mbox{\boldmath $\theta$}}
\newcommand{\bTheta}{\mbox{\boldmath $\Theta$}}
\newcommand{\blam}{\mbox{\boldmath $\Lambda$}}
\newcommand{\bdelta}{\mbox{\boldmath $\delta$}}
\newcommand{\bgamma}{\mbox{\boldmath $\gamma$}}
\newcommand{\balpha}{\mbox{\boldmath $\alpha$}}
\newcommand{\bbeta}{\mbox{\boldmath $\beta$}}
\newcommand{\balphascript}{\mbox{\boldmath ${\scriptstyle \alpha}$}}
\newcommand{\bbetascript}{\mbox{\boldmath ${\scriptstyle \beta}$}}
\newcommand{\bLambda}{\mbox{\boldmath $\Lambda$}}
\newcommand{\bDelta}{\mbox{\boldmath $\Delta$}}
\newcommand{\bomega}{\mbox{\boldmath $\omega$}}
\newcommand{\bOmega}{\mbox{\boldmath $\Omega$}}
\newcommand{\blambda}{\mbox{\boldmath $\lambda$}}
\newcommand{\bphi}{\mbox{\boldmath $\phi$}}
\newcommand{\bpi}{\mbox{\boldmath $\pi$}}
\newcommand{\bnu}{\mbox{\boldmath $\nu$}}
\newcommand{\brho}{\mbox{\boldmath $\rho$}}
\newcommand{\bmu}{\mbox{\boldmath $\mu$}}
\newcommand{\sigi}{\mbox{\boldmath $\Sigma$}_i}
\newcommand{\bfu}{{\bf u}}
\newcommand{\bfx}{{\bf x}}
\newcommand{\bfb}{{\bf b}}
\newcommand{\bfk}{{\bf k}}
\newcommand{\bfc}{{\bf c}}
\newcommand{\bfv}{{\bf v}}
\newcommand{\bfn}{{\bf n}}
\newcommand{\bfK}{{\bf K}}
\newcommand{\bfh}{{\bf h}}
\newcommand{\bff}{{\bf f}}
\newcommand{\bfg}{{\bf g}}
\newcommand{\bfe}{{\bf e}}
\newcommand{\bfr}{{\bf r}}
\newcommand{\bfw}{{\bf w}}
\newcommand{\calX}{{\cal X}}
\newcommand{\calZ}{{\cal Z}}
\newcommand{\bx}{{\bf x}}
\newcommand{\bb}{{\bf b}}
\newcommand{\by}{{\bf y}}
\newcommand{\bfy}{{\bf y}}
\newcommand{\bfz}{{\bf z}}
\newcommand{\bfs}{{\bf s}}
\newcommand{\bfa}{{\bf a}}
\newcommand{\bfA}{{\bf A}}
\newcommand{\bfB}{{\bf B}}
\newcommand{\bfV}{{\bf V}}
\newcommand{\bfZ}{{\bf Z}}
\newcommand{\bfH}{{\bf H}}
\newcommand{\bfX}{{\bf X}}
\newcommand{\bfR}{{\bf R}}
\newcommand{\bfF}{{\bf F}}
\newcommand{\bfS}{{\bf S}}
\newcommand{\bfC}{{\bf C}}
\newcommand{\bfI}{{\bf I}}
\newcommand{\bfO}{{\bf O}}
\newcommand{\bfU}{{\bf U}}
\newcommand{\bfD}{{\bf D}}
\newcommand{\bfY}{{\bf Y}}
\newcommand{\bSig}{{\bf \Sigma}}
\newcommand{\test}{\stackrel{<}{>}}
\newcommand{\zmk}{{\bf Z}_{m,k}}
\newcommand{\zlk}{{\bf Z}_{l,k}}
\newcommand{\zm}{{\bf Z}_{m}}
\newcommand{\ssq}{\sigma^{2}}
\newcommand{\dint}{{\displaystyle \int}}
\newcommand{\ds}{\displaystyle }
\newtheorem{theorem}{Theorem}
\newcommand{\postscript}[2]{ \begin{center}
    \includegraphics*[width=3.5in,height=#1]{#2.eps}
    \end{center} }

\newtheorem{identity}{Identity}
\newtheorem{hypothesis}{Hypothesis}
\newcommand{\mathtiny}[1]{\mbox{\tiny$#1$}}

\maketitle

\begin{abstract}
Normalizing flows (NF) recently gained attention as a way to construct
generative networks with exact likelihood calculation out of composable layers.
However, NF is restricted to dimension-preserving transformations.
Surjection VAE (SurVAE) has been proposed to extend NF to dimension-altering 
transformations.  Such networks are desirable because they are
expressive and can be precisely trained.  We show that the approaches 
are a re-invention of PDF projection, which appeared over twenty years earlier
and is much further developed.
\end{abstract}

\section{Introduction}
There is a need for expressive generative networks with 
exact likelihood calculation. To make this practical, such networks should be composable, 
representing complex operations as a cascade of simple layers,
and accumulating the likelihood contribution of the layers. 
The method of probability density function (PDF) Projection proposed as
early as 2000 \cite{Bagicpr,BagPDFProj}
provides an exact likelihood calculation for dimension-preserving
or dimension-reducing layers (we speak of the inference direction, i.e.
working from observed data to latent variables).
The approach can be applied recursively, so can accomodate composable
layer-based architectures.   A version of
PDF projection, applied to feed-forward neural networks (FFNN), called
projected belief network (PBN) was proposed in 2018\cite{BagPBN}.

Sadly, the method has been largely ignored by the machine learning community.
Recently, normalizing flows (NF) \cite{kobyzev2021normalizing},
which is limited to dimension-preserving transformations
has gained popularity, although the same idea of composable dimension-preserving
layers was proposed already in 2000 \cite{Bagicpr,BagPDFProj}.
To get around the restriction of dimension-preserving transformations,
surjection VAE (SurVAE) \cite{Nielsen_SurVAE_NEURIPS2020} has been proposed.
SurVAE provides exact likelihood calculations for dimension-reducing
transformations (inference surjections). SurVAE
also treats the problems of dimension-increasing,
stochastic or deterministic transformations, but
it is the deterministic dimension-reducing transformations that are 
the most interesting because these networks
form the bulk of machine learning (ML) and create generative models 
that concentrate information in lower-dimensional
output features. This can be seen as a general approach to
information maximization in networks \cite{BagKayInfo2022}.
In this paper, we show the equivalence of SurVAE and PDF projection 
using examples from  the SurVAE paper \cite{Nielsen_SurVAE_NEURIPS2020}.
With twenty years of development,
PDF projection has tackled a wider range of transformations.
We demonstrate the shortcomings of SurVAE in comparison to PDF projection,
mosly due to the different approaches to the problem.

\section{PDF Projection and PBN}
Consider a fixed and differentiable dimension-reducing transformation, $\bfz=T(\bfx)$, 
where $\bfx \in \mathbb{X}\subseteq \mathbb{R}^N$, and $\bfz \in \mathbb{R}^M$, where $M<N$.
Subject to mild constraints \cite{BagPDFProj,BagMaxEnt2018,Bag_info}, and assuming a known or assumed
feature distribution $g(\bfz)$, one can construct a probability density function (PDF)
on the input data with support $\mathbb{X}$ given by
\beq
G(\bfx) = \frac{p_{0,x}(\bfx)}{p_{0,x}(\bfz)} g(\bfz),    \;  \;  \; \bfz=T(\bfx),
\label{ppt0}
\eeq
where $p_{0,x}(\bfx)$ is a prior distribution and $p_{0,x}(\bfz)$
is its mapping to $\bfz$ through $T(\bfx)$. 
Note that (\ref{ppt0}) is a function of only $\bfx$ since $\bfz$ is 
deterministically determined from $\bfx$.
In our notation, the argument of the distribution defines its
range of support, and the variable in the subscript defines the original
range where the distribution was defined.
Thus, $p_{0,x}(\bfz)$ is a distribution over the range of $\bfz$, but
is a mapping of a distribution that was defined on $\bfx$.
It can be shown \cite{BagPDFProj,Bag_info} that 
$G(\bfx)$ is a PDF (integrates to 1) and is a member of the set of PDFs
that map to $g(\bfz)$ through $T(\bfx)$.
If $p_{0,x}(\bfx)$ is selected for maximum entropy (MaxEnt), 
then $G(\bfx)$ is unique for a given
transformation, data range $\mathbb{X}$, and a given $g(\bfz)$ (where ``g" represents
the ``given" feature distribution) \cite{Bag_info,BagMaxEnt2018}.  
To train the transformation, one maximizes the mean of
$\log G(\bfx)$ over a set of training data, and this results in a transformation
that extracts sufficient statistics and maximizes information \cite{BagKayInfo2022}.
We say that $G(\bfx)$ is the ``projection" of $g(\bfz)$ back to the input data
range $\mathbb{X}$, i.e. a {\it back-projection}.
We call the term $J(\bfx)\defined {p_{0,x}(\bfx) \over p_{0,x}(\bfz)}$ the ``J-function"
because in the special case of  dimension-preserving transformations
(i.e. normalizing flows \cite{kobyzev2021normalizing}),
$J(\bfx)$ is the determinant of the Jacobian of $T(\bfx)$.  
We show that  $\log J(\bfx)$ is identical to the ``likelihood contribution"
of SurVAE found in \cite{Nielsen_SurVAE_NEURIPS2020}.

Note that (\ref{ppt0}) can be seen as an application 
of Bayes rule \cite{Bag_info}  $G(\bfx)=p_{0,x}(\bfx|\bfz) g(\bfz)$,
where  $p_{0,x}(\bfx|\bfz)$ is the {\it a posteriori} distribution
under prior $p_{0,x}(\bfx)$, also equal to the J-function, i.e.
\beq
J(\bfx) = \frac{p_{0,x}(\bfx)}{p_{0,x}(\bfz)} = p_{0,x}(\bfx|\bfz).
\label{jfpost}
\eeq
Because $T(\bfx)$ is deterministic, the posterior distribution
$p_{0,x}(\bfx|\bfz)$ has support only on a set of zero volume, the level set 
(called  {\it fiber of }$\bfz$ in \cite{Nielsen_SurVAE_NEURIPS2020}) 
\beq
  {\cal M}(\bfz) = \{ \bfx \in \mathbb{X} | T(\bfx) = \bfz\},
  \label{mandef}
\eeq
To show that (\ref{jfpost}) integrates to 1 on ${\cal M}(\bfz)$,
we integrate the J-function for fixed $\bfz$:
\beq
\int_{\bfx\in{\cal M}(\bfz)} \; \frac{p_{0,x}(\bfx)}{p_{0,x}(\bfz)}  \; {\rm d} \bfx = \frac{1}{p_{0,x}(\bfz)}  \; \int_{\bfx\in{\cal M}(\bfz)} \; p_{0,x}(\bfx) \; {\rm d} \bfx  = 1.
\label{jfint}
\eeq
We used that the denominator of (\ref{jfint}) is independent 
of $\bfx$ on ${\cal M}(\bfz)$ and the identity 
$\int_{\bfx\in{\cal M}(\bfz)} \; p_{0,x}(\bfx) \; {\rm d} \bfx  = p_{0,x}(\bfz).$

To generate data from $G(\bfx)$ in (\ref{ppt0}), one
draws a sample $\bfz$ from $g(\bfz)$, then draws a sample $\bfx$ from  
 ${\cal M}(\bfz)$ and weighted by the prior distribution $p_{0,x}(\bfx)$.
In other words, we sample from $p_{0,x}(\bfx)$ restricted 
and normalized on ${\cal M}(\bfz)$.

%
For complex transformations, where it is
impossible to find a tractable form for $p_{0,x}(\bfz)$,
one breaks the transformation into a cascade of simpler transformations
and uses the chain-rule.  Consider a cascade of two transformations, $\bfy=T_1(\bfx)$, and $\bfz=T_2(\bfy)$.
Then, applying (\ref{ppt0}) recursively,
\beq
G(\bfx) = \frac{p_{0,x}(\bfx)}{p_{0,x}(\bfy)} \frac{p_{0,y}(\bfy)}{p_{0,y}(\bfz)} g(\bfz),
\label{ppt1}
\eeq
which can be extended to any number of stages.
To compute $\log G(\bfx)$, one just accumulates the log-J function of the transformations.
Data generation is also cascaded, and is initiated by drawing a sample $\bfz$ from $g(\bfz)$.

When PDF projection is applied to a feed-forward neural network (FFNN) layer-by-layer, this results in the
projected belief network (PBN) \cite{BagPBN}.

\section{SurVAE}
In this section, we briefly describe SurVAE as we compare it mathematically to PDF projection,
limiting ourselves, for reasons already discussed, to what the authors of SurVAE call deterministic 
{\it inference surjections}, which are dimension-reducing deterministic transformations.
For an inference surjections, the likelihood of $\bfx$ is given by  \cite{Nielsen_SurVAE_NEURIPS2020}
\beq
   \log p(\bfx) = \mathbb{E}_{q(\bfz|\bfx)}\left[ \log p(\bfz) \right] + \mathbb{E}_{q(\bfz|\bfx)}\left[ \log \frac{p(\bfx|\bfz)}{q(\bfz|\bfx)}\right],
\eeq
where $q(\bfz|\bfx)$ is the distribution of $\bfz$ given $\bfx$, 
simplifying for a deterministic transformation to $q(\bfz|\bfx)=\delta[\bfz-T(\bfx)],$
so  the first term becomes 
$\mathbb{E}_{q(\bfz|\bfx)}\left[ \log p(\bfz) \right] = \log p(\bfz).$
In the second term, which is the {\it likelihood contribution},
both numerator and denominator have a term $\delta[\bfz-T(\bfx)]$,
so it can be reduced to ${\cal V}(\bfx,\bfz) =   \log p(\bfx|\bfz)$,
however this is the same as the J-function as seen by (\ref{jfpost}),
but instead of being based on a prior, the posterior
$p(\bfx|\bfz)$ is directly specified.

\section{Comparison of PDF projection and SurVAE}
The difference in the approaches can bs concicely stated as follows.
In SurVAE, it is necessary to specify $p(\bfx|\bfz)$,
then find the forward transformation $T(\bfx)$, which they call ``inverse" transformation,
because it inverts the dimension-increasing transformation implied by $p(\bfx|\bfz)$.
In contrast, in PDF projection, $p_{0,x}(\bfx)$ and $T(\bfx)$ are directly
specified, then $p_{0,x}(\bfz)$ must be derived.  Which approach is best may depend 
on the application.  Note that the existence of a prior distribution for SurVAE
is implied by the choice of $p(\bfx|\bfz)$, but is not
explicitly stated.  In SurVAE, specifying $p(\bfx|\bfz)$ may not be trivial, 
since it must have an  ``inverse function" $T(\bfx)$
and must have support only on (\ref{mandef}).
We now show the equivalence using two examples provided in \cite{Nielsen_SurVAE_NEURIPS2020}.  We then show a third example where the solution is more apparent 
using PDF projection.

\subsection{Tensor Slicing}
In tensor slicing, the output vector $\bfz$ consists of a subset
of the elements of $\bfx$.
Let $\bfz=\{x_1,x_2 \ldots x_M\}.$ Then, if we use the zero-mean variance 1 Gaussian prior
$\log p_{x,0}(\bfx) = -\frac{N}{2} \log (2 \pi) - \frac{1}{2}\sum_{i=1}^N x_i^2,$ 
and
$\log p_{x,0}(\bfz) = -\frac{M}{2} \log (2 \pi) - \frac{1}{2}\sum_{i=1}^M x_i^2,$
so
$\log J(\bfx) = \log \frac{p_{x,0}(\bfx)}{p_{x,0}(\bfz)}=
-\frac{N-M}{2} \log (2 \pi)   - \frac{1}{2}\sum_{i=M+1}^N x_i^2.$
In \cite{Nielsen_SurVAE_NEURIPS2020}, Example 1, their result
$-\log q(\bfz_2|\bfx)$ is the same, 
but is negated because the definitions of $\bfx$ and $\bfz$ are switched.

\subsection{Absolute Value}
Consider the absolute value operator
$z_i=|x_i|$, $i=1,2 \ldots N$.
If we use the Gaussian prior (same as above example), then $p_{x,0}(\bfz)$
is the truncated Gaussian, truncated to $[0, \infty]$,
so equals $p_{x,0}(\bfz) = 2^N p_{x,0}(\bfx)$. Therefore,
$\log p_{x,0}(\bfz) = N \log 2 + \log p_{x,0}(\bfx),$
so $\log J(\bfx) = - N \log 2,$ which agrees with the 
result in \cite{Nielsen_SurVAE_NEURIPS2020}, 
in Section 3.1, ``Abs Surjection", when their result is
simplified for a deterministic transformation and a symmetric prior 
(probability of a plus and minus sign are both 0.5), then the result is $(0.5)^N$,
which is the same.

\subsection{Linear transforms of bounded data}
\label{spasec}
Consider the linear dimension-reducing transformation
$\bfz={\bf W}^\prime \bfx$, where ${\bf W}$ is an $N\times M$ full-rank matrix,
where $M<N$ and the elements of $\bfx$ are bounded in $[0, 1]$.
This is a typical fully-connected neural network layer with bounded input data.
In keeping with maximum entropy, we choose the uniform prior 
$p_{0,x}(\bfx)=1$. The J-function (i.e. likelihood contribution) is then given by
$J(\bfx)=\frac{1}{p_{0,x}(\bfz)}$. Except in simple cases
(i.e. Irwin-Hall distribution), a linear transformation of uniform random
variables produces an intractable distribution.
However, Steven Kay noted that in cases like this, the moment generating function (MGF) can be
written down in closed form, and this could be inverted 
using the saddle point approximation (SPA) to find $p_{0,x}(\bfz)$ \cite{BagNutKay2000}.
The inversion is based on approximating the MGF along a path of integration
at the saddle point, which due to central limit theorem, quickly approaches Gaussian.
Errors in the inversion of the MGF are almost negligible \cite{BagSPL2021}
because they result only from the mismatch of the shape of the 
integrand, not the actual values. The method is accurate in the far tails of the 
distribution, something that one would not expect from an ``approximation". 
Details of the SPA for the uniform prior can be found in
the appendix of \cite{BagEusipcoRBM}.  The approach can be extended to other priors
in the data in the range $[0, \infty]$, such as the exponential prior (See \cite{BagUMS}), or the
truncated Gaussian prior distribution \cite{BagIcasspPBN}.  

\section{Conclusions}
SurVAE and PDF Projection are two significantly different approaches to 
the same method.  In SurVAE, a wider range of transformations
are considered, dimension increasing, dimension reducing, deterministic, stochastic.
But for the more important deterministic dimension-reducing case, 
PDF projection has a richer choice of transformations,
many provided in a tutorial paper \cite{BagAESTut},
with implementations available online \cite{PBNTk}.
Even for simple linear transformations of bounded data
provided in Section \ref{spasec}, the SurVAE approach is intractable,  but using PDF projection,
the MGF of $p_{x,0}(\bfz)$ is available
in closed form, so can be solved with the SPA.

\bibliographystyle{IEEEtran}
\bibliography{ppt}
\end{document}